\pdfoutput=1
\documentclass{article}

\newif\ifarxiv
\arxivtrue
\ifarxiv
  \usepackage[preprint]{neurips_2025}
\else
  \usepackage[dblblindworkshop]{neurips_2025}
\fi

\usepackage[utf8]{inputenc}
\usepackage[T1]{fontenc}
\usepackage{hyperref}
\usepackage{url}
\usepackage{booktabs}
\usepackage{amsfonts}
\usepackage{amsmath}
\usepackage{amssymb}
\usepackage{nicefrac}
\usepackage{microtype}
\usepackage{graphicx}
\usepackage{multirow}
\usepackage{xcolor}
\usepackage{tikz}
\usepackage{enumitem}
\usetikzlibrary{shapes.geometric, arrows.meta, positioning}
\definecolor{tgreen}{RGB}{190,214,165}
\definecolor{eblue}{RGB}{176,206,222}

\newcommand{\todo}[1]{}   %
\newcommand{\num}[1]{#1}   %

\makeatletter
\ifarxiv
\renewcommand{\@noticestring}{NeurIPS 2026 Workshop on Principles of Generative Modeling.}
\else
\renewcommand{\@noticestring}{Submitted to the NeurIPS 2026 Workshop on Principles of Generative Modeling (PriGM). Do not distribute.}
\fi
\makeatother

\title{Diffusable Latents from Structure-Agnostic Distillation}

\workshoptitle{Principles of Generative Modeling (PriGM) @ NeurIPS 2026}

\ifarxiv
\author{%
  Adrien Ramanana Rahary$^{1,2}$ \quad Nicolas Dufour$^{2}$ \quad Patrick P\'erez$^{2}$ \quad David Picard$^{1}$\\[3pt]
  $^{1}$LIGM, ENPC, IP Paris, CNRS, UGE \qquad $^{2}$Kyutai\\[2pt]
  \texttt{\{adrien.ramanana-rahary,\,david.picard\}@enpc.fr}\\
  \texttt{\{nicolas.dufour,\,patrick\}@kyutai.org}
}
\else
\author{%
  Anonymous Author(s)\\
  Affiliation\\
  Address\\
  \texttt{email}\\
}
\fi

\begin{document}
\maketitle

\begin{abstract}
Distilling pretrained foundation models into an autoencoder bottleneck improves latent
diffusability, enabling diffusion models to converge faster and reach higher sample quality.
Standard distillation aligns the latent at each position to a co-located teacher feature, tying the
latent layout to the teacher's. We show this constraint is unnecessary: aligning a single pooled
image-level descriptor to the teacher's performs as well as or slightly better than
dense position-wise distillation. We compare first-order and relational pooled objectives across
latent shapes and teacher modalities. First-order matching extends naturally to 1D token-sequence
latents and across modalities, where distilling a text encoder into an image autoencoder still
improves diffusability; a relational objective based only on each image's nearest neighbours
improves it as well. Code and blog post are available at
\href{https://github.com/AdrienRR/structure-agnostic-distillation}{\nolinkurl{github.com/AdrienRR/structure-agnostic-distillation}} and
\href{https://kyutai.org/blog/2026-09-28-structure-agnostic-distillation/}{\nolinkurl{kyutai.org/blog/2026-09-28-structure-agnostic-distillation}}.
\end{abstract}

\section{Introduction}
\label{sec:intro}
Latent diffusion trains a diffusion prior in an autoencoder's latent
\citep{rombach2022ldm, peebles2023dit}, and sample quality depends on how
diffusable that latent is \citep{skorokhodov2025diffusability}. Prior work makes the latent more diffusable by distilling a frozen pretrained
vision encoder into the autoencoder bottleneck, matching each latent position to its
co-located patch feature
\citep{peng2022beitv2, hu2023gaia1, russell2025gaia2, yao2025vavae, bi2026vfmvae}. A more extreme line replaces the trained encoder with a frozen representation model and diffuses in its output space \citep{zheng2025rae, tong2026scalerae,
singh2026raev2, mira2026, guo2026vrae}. Both approaches place the latent and the teacher on a shared patch grid.
But such a grid can be unavailable: compact 1D token
latents, a short list of tokens with no spatial layout \citep{yu2024titok,
bachmann2025flextok}, drop it on the student side for efficiency, and a teacher from
another modality, such as a text encoder, never had one to share. Position-wise
distillation then reaches these settings only through an artificial correspondence,
as SoftVQ-VAE \citep{chen2025softvq} does by replicating latent tokens and learning
a projector to a pretrained grid. We ask whether distillation can improve diffusability
without any token-level correspondence between the latent and the teacher.

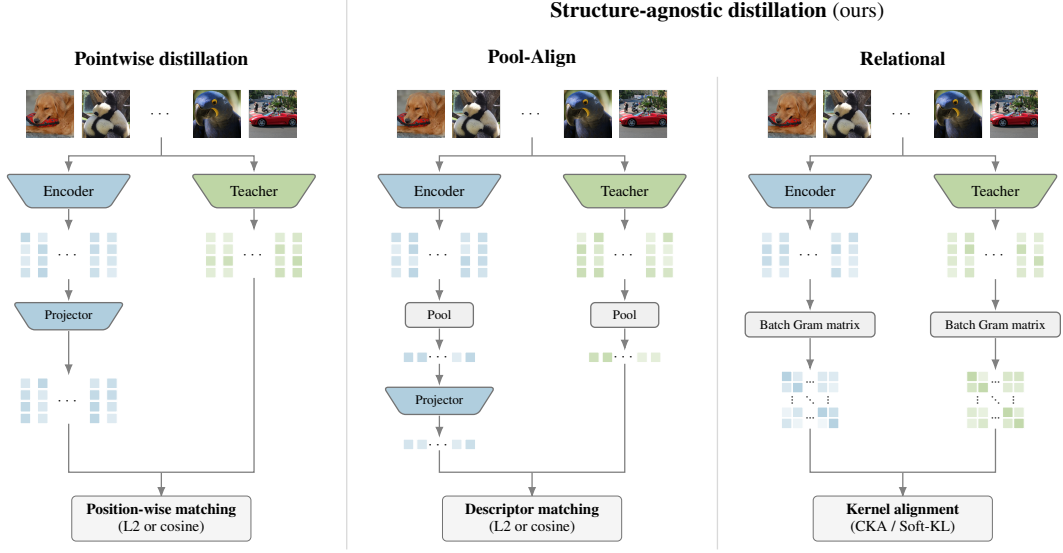
\begin{figure}[t]
\centering
\pgfmathsetseed{7}
\resizebox{\linewidth}{!}{%
\begin{tikzpicture}[
  algn/.style={rounded corners=2pt, draw=black!55, fill=black!4, minimum width=2.9cm, minimum height=0.66cm, align=center, font=\scriptsize},
  ar/.style={-{Latex[length=1.8mm]}, black!50, semithick},
  ln/.style={black!50, semithick},
  cell/.style={draw=white, line width=0.35pt, minimum size=1.7mm, inner sep=0pt},
  im/.style={inner sep=0, draw=gray!45, line width=0.3pt},
]
\def\PG#1#2#3{%
  \foreach \c in {0,1}{\foreach \r in {0,1,2,3}{\pgfmathsetmacro{\pp}{int(35+50*rnd)}%
    \node[cell,fill=#3!\pp] at (#1-0.7+\c*0.28,#2-\r*0.19){};}}%
  \foreach \c in {0,1}{\foreach \r in {0,1,2,3}{\pgfmathsetmacro{\pp}{int(35+50*rnd)}%
    \node[cell,fill=#3!\pp] at (#1+0.42+\c*0.28,#2-\r*0.19){};}}%
  \node[font=\tiny] at (#1,#2-0.29){$\cdots$};}
\def\KR#1#2#3{\foreach \r/\ro in {0/-0.38,1/-0.20,3/0.20,4/0.38}{\foreach \c/\co in {0/-0.38,1/-0.20,3/0.20,4/0.38}{%
  \pgfmathsetmacro{\kx}{#1+\co}\pgfmathsetmacro{\ky}{#2-(0.38+\ro)}%
  \pgfmathparse{\r==\c?92:int(18+52*rnd)}\edef\pp{\pgfmathresult}\node[cell,fill=#3!\pp] at (\kx,\ky){};}}%
  \pgfmathsetmacro{\kxc}{#1}\pgfmathsetmacro{\kyc}{#2-0.38}%
  \pgfmathsetmacro{\kya}{#2-0.09}\pgfmathsetmacro{\kyb}{#2-0.67}%
  \pgfmathsetmacro{\kxa}{#1-0.29}\pgfmathsetmacro{\kxb}{#1+0.29}%
  \fill[black!65] (\kxc-0.05,\kya) circle (0.4pt) (\kxc,\kya) circle (0.4pt) (\kxc+0.05,\kya) circle (0.4pt);%
  \fill[black!65] (\kxc-0.05,\kyb) circle (0.4pt) (\kxc,\kyb) circle (0.4pt) (\kxc+0.05,\kyb) circle (0.4pt);%
  \fill[black!65] (\kxa,\kyc+0.05) circle (0.4pt) (\kxa,\kyc) circle (0.4pt) (\kxa,\kyc-0.05) circle (0.4pt);%
  \fill[black!65] (\kxb,\kyc+0.05) circle (0.4pt) (\kxb,\kyc) circle (0.4pt) (\kxb,\kyc-0.05) circle (0.4pt);%
  \fill[black!65] (\kxc-0.05,\kyc+0.05) circle (0.4pt) (\kxc,\kyc) circle (0.4pt) (\kxc+0.05,\kyc-0.05) circle (0.4pt);}
\def\TRAP#1#2#3#4{\draw[fill=#3,draw=black!55,semithick,rounded corners=2pt]%
  (#1-1.02,#2+0.27)--(#1+1.02,#2+0.27)--(#1+0.66,#2-0.27)--(#1-0.66,#2-0.27)--cycle;%
  \node[font=\scriptsize] at (#1,#2){#4};}
\def\PROJ#1#2{\draw[fill=eblue,draw=black!55,semithick,rounded corners=2pt]%
  (#1-0.9,#2+0.22)--(#1+0.9,#2+0.22)--(#1+0.62,#2-0.22)--(#1-0.62,#2-0.22)--cycle;%
  \node[font=\tiny] at (#1,#2){Projector};}
\def\POOL#1#2{\draw[fill=black!6,draw=black!55,semithick,rounded corners=2pt]%
  (#1-0.55,#2+0.2) rectangle (#1+0.55,#2-0.2);\node[font=\tiny] at (#1,#2){Pool};}
\def\GRAM#1#2{\draw[fill=black!6,draw=black!55,semithick,rounded corners=2pt]%
  (#1-1.05,#2+0.2) rectangle (#1+1.05,#2-0.2);\node[font=\tiny] at (#1,#2){Batch Gram matrix};}
\def\PD#1#2#3{%
  \foreach \c in {0,1}{\pgfmathsetmacro{\pp}{int(35+50*rnd)}\node[cell,fill=#3!\pp] at (#1-0.5+\c*0.22,#2){};}%
  \node[font=\tiny] at (#1,#2){$\cdots$};%
  \foreach \c in {0,1}{\pgfmathsetmacro{\pp}{int(35+50*rnd)}\node[cell,fill=#3!\pp] at (#1+0.28+\c*0.22,#2){};}}
\def\IMGS#1{\node[im] at (#1-1.8,6.0){\includegraphics[width=7.6mm]{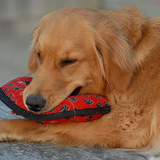}};
  \node[im] at (#1-0.9,6.0){\includegraphics[width=7.6mm]{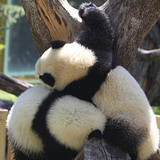}};
  \node[font=\tiny] at (#1,6.0){$\cdots$};
  \node[im] at (#1+0.9,6.0){\includegraphics[width=7.6mm]{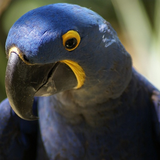}};
  \node[im] at (#1+1.8,6.0){\includegraphics[width=7.6mm]{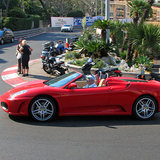}};}
\def\WIRE#1{\draw[ln](#1,5.62)--(#1,5.32);%
  \draw[ar](#1,5.32)-|(#1-1.5,5.05);\draw[ar](#1,5.32)-|(#1+1.5,5.05);}
\def\CX{0}
\node[font=\small\bfseries] at (\CX,6.9){Pointwise distillation};
\IMGS{\CX}
\TRAP{\CX-1.5}{4.75}{eblue}{Encoder}\TRAP{\CX+1.5}{4.75}{tgreen}{Teacher}
\WIRE{\CX}
\PG{\CX-1.5}{4.0}{eblue}\PG{\CX+1.5}{4.0}{tgreen}
\draw[ar](\CX-1.5,4.46)--(\CX-1.5,4.105);\draw[ar](\CX+1.5,4.46)--(\CX+1.5,4.105);
\PROJ{\CX-1.5}{2.725}
\draw[ar](\CX-1.5,3.325)--(\CX-1.5,2.965);
\PG{\CX-1.5}{1.64}{eblue}
\draw[ar](\CX-1.5,2.485)--(\CX-1.5,1.765);
\draw[ln](\CX-1.5,0.985)|-(\CX,0.2);
\draw[ln](\CX+1.5,3.345)|-(\CX,0.2);
\draw[ar](\CX,0.2)--(\CX,-0.22);
\node[algn](Lal) at (\CX,-0.55){\textbf{Position-wise matching}\\(L2 or cosine)};
\def\MX{6.0}
\node[font=\small\bfseries] at (\MX,6.9){Pool-Align};
\IMGS{\MX}
\TRAP{\MX-1.5}{4.75}{eblue}{Encoder}\TRAP{\MX+1.5}{4.75}{tgreen}{Teacher}
\WIRE{\MX}
\PG{\MX-1.5}{4.0}{eblue}\PG{\MX+1.5}{4.0}{tgreen}
\draw[ar](\MX-1.5,4.46)--(\MX-1.5,4.105);\draw[ar](\MX+1.5,4.46)--(\MX+1.5,4.105);
\POOL{\MX-1.5}{2.745}\POOL{\MX+1.5}{2.745}
\draw[ar](\MX-1.5,3.325)--(\MX-1.5,2.965);\draw[ar](\MX+1.5,3.325)--(\MX+1.5,2.965);
\PD{\MX-1.5}{2.06}{eblue}\PD{\MX+1.5}{2.06}{tgreen}
\draw[ar](\MX-1.5,2.525)--(\MX-1.5,2.165);\draw[ar](\MX+1.5,2.525)--(\MX+1.5,2.165);
\PROJ{\MX-1.5}{1.355}
\draw[ar](\MX-1.5,1.955)--(\MX-1.5,1.595);
\PD{\MX-1.5}{0.65}{eblue}
\draw[ar](\MX-1.5,1.115)--(\MX-1.5,0.755);
\draw[ln](\MX-1.5,0.555)|-(\MX,0.2);
\draw[ln](\MX+1.5,1.955)|-(\MX,0.2);
\draw[ar](\MX,0.2)--(\MX,-0.22);
\node[algn](Mal) at (\MX,-0.55){\textbf{Descriptor matching}\\(L2 or cosine)};
\def\DX{12.0}
\node[font=\small\bfseries] at (\DX,6.9){Relational};
\IMGS{\DX}
\TRAP{\DX-1.5}{4.75}{eblue}{Encoder}\TRAP{\DX+1.5}{4.75}{tgreen}{Teacher}
\WIRE{\DX}
\PG{\DX-1.5}{4.0}{eblue}\PG{\DX+1.5}{4.0}{tgreen}
\draw[ar](\DX-1.5,4.46)--(\DX-1.5,4.105);\draw[ar](\DX+1.5,4.46)--(\DX+1.5,4.105);
\GRAM{\DX-1.5}{2.59}\GRAM{\DX+1.5}{2.59}
\draw[ar](\DX-1.5,3.325)--(\DX-1.5,2.81);\draw[ar](\DX+1.5,3.325)--(\DX+1.5,2.81);
\KR{\DX-1.5}{1.75}{eblue}\KR{\DX+1.5}{1.75}{tgreen}
\draw[ar](\DX-1.5,2.37)--(\DX-1.5,1.845);\draw[ar](\DX+1.5,2.37)--(\DX+1.5,1.845);
\draw[ln](\DX-1.5,0.905)|-(\DX,0.2);\draw[ln](\DX+1.5,0.905)|-(\DX,0.2);
\draw[ar](\DX,0.2)--(\DX,-0.22);
\node[algn](Ral) at (\DX,-0.55){\textbf{Kernel alignment}\\(CKA / Soft-KL)};
\draw[black!15,semithick](3.0,-1.05)--(3.0,7.85);
\node[font=\bfseries] at (9.0,7.65){Structure-agnostic distillation {\normalfont(ours)}};
\draw[black!15,semithick](9.0,-1.05)--(9.0,6.6);
\end{tikzpicture}}
\caption{\textbf{Pointwise vs.\ structure-agnostic distillation of a foundation encoder into the
autoencoder latent.} A batch of $B$ images is encoded by the frozen teacher and the trainable
encoder. \textbf{Pointwise} (left) matches each latent token to its co-located teacher token,
requiring a shared token grid and a channel projector. \textbf{Pool-Align} (middle) matches one
pooled descriptor per image, requiring only a projector, not a grid. \textbf{Relational} (right)
matches the $B\times B$ between-image similarity kernel, requiring neither. The structure-agnostic
variants thus apply to any latent and teacher shape.}
\label{fig:teaser}
\end{figure}

We study \textbf{structure-agnostic distillation}: we pool each image's latent
tokens into one image-level descriptor and define the distillation
objective on these descriptors alone, leaving the latent's internal layout free
(Figure~\ref{fig:teaser}). Because pooling reduces each image to a single vector,
the objective is independent of the latent's token count and layout, so student and
teacher may take any shape. We compare a family of
pooled objectives: a \emph{first-order} match that aligns each descriptor directly
to its teacher descriptor, and \emph{relational} objectives \citep{park2019rkd,
passalis2018pkt, tung2019similarity, tian2020crd} that instead match the
matrix of between-image similarities, via centered kernel alignment
(\emph{CKA}, \citealt{kornblith2019cka}) or a softmaxed-similarity KL divergence
(\emph{Soft-KL}); both are detailed in Section~\ref{sec:method}.

To our knowledge, this is the first study to show that aligning pooled representations can replace position-wise alignment while improving diffusability. Varying only the distillation loss at fixed
architecture and budget, first-order pooled matching performs as well as or slightly better than position-wise alignment where the latter applies, and structure-agnostic distillation extends to shapes that position-wise alignment reaches
only through an artificial correspondence, handling a 1D token-sequence latent and letting a
text encoder improve the diffusability of an image latent. Representation distillation can transfer
useful semantic geometry into a generative latent without inheriting the teacher's structural organization.

\section{Method}
\label{sec:method}

\paragraph{Preliminaries.} An autoencoder maps an image $x$ to a latent $z=E(x)$ with an
encoder $E$ and reconstructs it as $\hat x=D(z)$ with a decoder $D$, the two
trained jointly with $\mathcal{L}_{\text{vae}}$, the standard image-VAE objective
(pixel, perceptual, adversarial, and variational terms; Appendix~\ref{app:vaeloss}). A diffusion prior is
trained afterwards in this latent space. To make $z$ more diffusable,
we add an auxiliary term $\mathcal{L}_{\text{distill}}$ to the encoder's objective,
aligning the latent to a frozen foundation encoder. The encoder $E$
minimizes
$\mathcal{L}_{\text{vae}}+\lambda\,\mathcal{L}_{\text{distill}}$, where $\lambda$
weights the two terms. This section
develops $\mathcal{L}_{\text{distill}}$.

\paragraph{Pooled distillation objectives.} For image $i$ in a batch of $B$, each
structure-agnostic objective mean-pools the student's $N_s$ latent tokens
$z_i\in\mathbb{R}^{N_s\times d_s}$ (resp.\ the teacher's $N_t$ features
$g_i\in\mathbb{R}^{N_t\times d_t}$) into one $L_2$-normalized descriptor $\hat u_i\in\mathbb{R}^{d_s}$
(resp.\ $\hat v_i\in\mathbb{R}^{d_t}$; the two dimensions generally differ). Pooling discards the token layout, so the same objective works
whatever the latent's shape. They relax the correspondence position-wise
distillation requires: position-wise ties each latent token to its co-located teacher token
($z_{i,j}\!\leftrightarrow\! g_{i,j}$); a first-order pooled match ties one descriptor per image
($\hat u_i\leftrightarrow\hat v_i$); a relational match ties none, matching only the between-image
similarity matrices ($S_s\leftrightarrow S_t$, with $S_{s,ij}=\hat u_i^\top\hat u_j$ and $S_{t,ij}=\hat v_i^\top\hat v_j$). We instantiate
three: one first-order and two relational.
\begin{itemize}[leftmargin=1.3em, itemsep=3pt, topsep=3pt, parsep=0pt]
\item \textbf{Pool-Align}. A first-order match aligns each image's pooled student descriptor with
the teacher's descriptor for the same image. A learned linear map $W$ sends $\hat u_i$ into the
teacher's space, and the loss is the cosine distance
$\mathcal{L}_{\text{pool}} = \sum_i \bigl(1 - \hat v_i^\top \tilde u_i\bigr)$, with
$\tilde u_i = W\hat u_i/\lVert W\hat u_i\rVert$.
\item \textbf{Relational.} A relational objective reproduces the teacher's between-image
similarity structure, not the descriptors themselves. Both compare images only through
the $B\times B$ matrices $S_s,S_t$ above, built from $L_2$-normalized tokens; the matrix is
invariant to the descriptor space, so they need no projector.
\begin{itemize}[leftmargin=1.1em, itemsep=3pt, topsep=3pt, parsep=0pt, label=$\circ$]
\item \textbf{Centered kernel alignment} (CKA) reproduces the teacher's global geometry
(images placed close stay close), matching the two matrices as a whole,
$\mathcal{L}_{\text{CKA}} = 1 - \mathrm{CKA}(S_s,S_t)$, with the unbiased estimator
\citep{kornblith2019cka} (Appendix~\ref{app:cka}).
\item \textbf{Softmaxed-similarity KL} (Soft-KL) matches each image's soft ranking of its nearest neighbours: a
softmax with temperature $\tau$ turns each row of $S$ into a neighbour distribution,
matched by KL,
$p_{i} = \mathrm{softmax}_{j\neq i}(S_{ij}/\tau)$ and
$\mathcal{L}_{\text{softKL}} = \sum_i \mathrm{KL}(p^t_{i}\,\|\,p^s_{i})$.
\end{itemize}
\end{itemize}

\section{Experiments}
\label{sec:exp}
\paragraph{Setup.} We build on the VA-VAE implementation \citep{yao2025vavae}, adopting its
autoencoder and LightningDiT-XL prior. All arms use the same architecture and training recipe
(Appendix~\ref{app:arch});
only the distillation term that shapes the latent, or its absence,
varies. This term acts on the encoder's posterior mean (Appendix~\ref{app:lambda}). Each arm trains the autoencoder on
$256{\times}256$ ImageNet \citep{deng2009imagenet} from scratch (50 epochs, batch 256), then a
LightningDiT-XL prior (64 epochs, batch 1024). We compare
arms by gFID \citep{heusel2017fid} without classifier-free guidance (no per-arm
tuning); for the 2D grid we verify the ranking is preserved under VA-VAE's guidance
recipe (Appendix~\ref{app:arch}; samples in Appendix~\ref{app:samples}).

\paragraph{Study arms.} We compare our structure-agnostic methods against two
references: a control with no distillation (\emph{No Distillation}), and the
position-wise baseline, VA-VAE's \citep{yao2025vavae} vision-foundation alignment (\emph{VF}), which aligns
each latent cell to its co-located teacher feature.
Our family, defined in Section~\ref{sec:method}, spans the first-order
pooled alignment (\emph{Pool-Align}) and two relational objectives, centered kernel alignment
(\emph{CKA}) and softmaxed-similarity KL (\emph{Soft-KL}).

\paragraph{Settings.} We run these across three settings that vary the student's
shape or the teacher's modality:
\begin{enumerate}[label=\textbf{(\Alph*)}, leftmargin=2.1em, itemsep=1pt, topsep=1pt, parsep=0pt]
\item A \textbf{2D grid} latent with a DINOv2 image teacher \citep{oquab2023dinov2}, where the
latent grid lines up with the teacher's and all five methods apply, so this setting measures
the structure-agnostic methods against position-wise \emph{VF} on equal footing.
\item A \textbf{1D sequence} latent with the same image teacher, produced by a
Perceiver-resampler \citep{jaegle2021perceiver} bottleneck that emits a flat set of tokens with
no spatial grid; position-wise \emph{VF} then has no grid to align to and drops out, while the
four layout-free methods still apply.
\item A \textbf{cross-modal} 2D grid whose latent we align to per-image captions from the
captioned ImageNet of \citet{degeorge2025imagenett2i}, embedded with a BGE-large text teacher
\citep{xiao2023bge}, which has no spatial grid, so \emph{VF} drops out here too.
\end{enumerate}

\begin{table}[t]
\centering
\caption{Generation quality (gFID, $\downarrow$), without classifier-free guidance (no per-arm
tuning); for the 2D grid the ranking is preserved under VA-VAE's guidance recipe
(Appendix~\ref{app:arch}). Autoencoders are trained for $50$ epochs on $256\times256$ ImageNet and
the diffusion priors for $64$ epochs (Appendix~\ref{app:arch}). Rows are settings A--C (bottleneck
structure $\times$ teacher model). The structure-agnostic methods (\emph{Pool-Align}, \emph{CKA},
\emph{Soft-KL}) apply in every setting; the position-wise baseline \emph{VF} needs a shared
latent--teacher grid, so it is inapplicable (\emph{n/a}) in the 1D~(B) and cross-modal~(C)
settings. Within a row only the distillation method changes, with architecture and
budget fixed, so differences reflect the objective, not compute; absolute gFID differs across
bottleneck structures, so we compare only within a setting. Best per row in \textbf{bold}.}
\label{tab:gfid}
\resizebox{\linewidth}{!}{%
\begin{tabular}{lllccccc}
\toprule
 & & & \multicolumn{2}{c}{Baselines} & \multicolumn{3}{c}{Structure-agnostic (ours)} \\
\cmidrule(lr){4-5}\cmidrule(lr){6-8}
Setting & Bottleneck & Teacher & No Distill. & VF~\citep{yao2025vavae} & Pool-Align & CKA & Soft-KL \\
\midrule
A: shared grid & 2D grid & DINOv2~\citep{oquab2023dinov2}    & \num{9.52}  & \num{6.04} & \num{\textbf{5.77}}  & \num{7.06}  & \num{6.51}  \\
B: no grid     & 1D seq  & DINOv2~\citep{oquab2023dinov2}    & \num{26.49} & \emph{n/a}         & \num{\textbf{15.62}} & \num{18.13} & \num{17.87} \\
C: cross-modal & 2D grid & BGE-large~\citep{xiao2023bge}     & \num{9.52}  & \emph{n/a}         & \num{\textbf{6.97}} & \num{10.41} & \num{8.51}  \\
\bottomrule
\end{tabular}%
}
\end{table}

\begin{figure}[t]
\centering
\includegraphics[width=\linewidth]{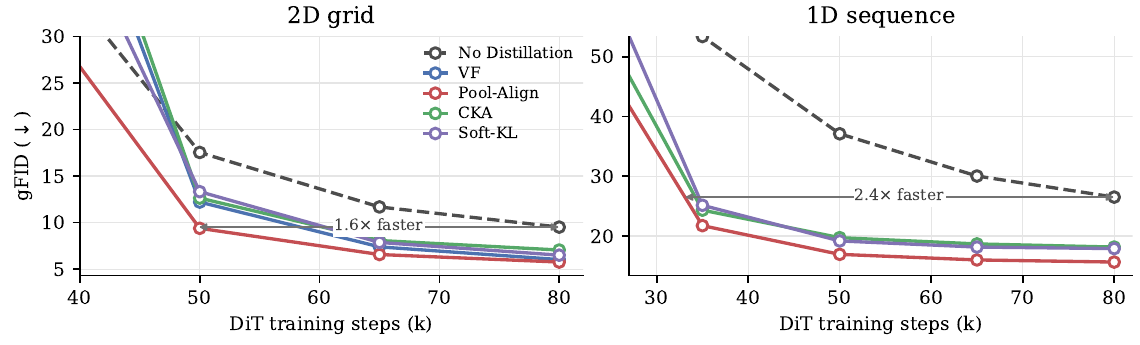}
\caption{\textbf{Distillation speeds up diffusion training.} Per-checkpoint gFID ($50$k samples,
without classifier-free guidance) against DiT training steps for the 2D-grid (left) and
1D-sequence (right) settings. Every distilled latent converges to a lower final gFID than
\emph{No-Distillation} and reaches low gFID in fewer steps;
the horizontal marker reports how many fewer steps \emph{Pool-Align} needs to reach
\emph{No-Distillation}'s final gFID.}
\label{fig:convergence}
\end{figure}

\paragraph{Results.}
Table~\ref{tab:gfid} reports final gFID and Figure~\ref{fig:convergence} the in-modality
training-time convergence.
\begin{enumerate}[label=\textbf{(\Alph*)}, leftmargin=1.9em, itemsep=3pt, topsep=3pt, parsep=0pt]
\item \emph{Shared grid}: does discarding token correspondence hurt? No. Pooling performs as well as or slightly better than position-wise \emph{VF} (\emph{Pool-Align} \num{5.77} vs.\ \emph{VF}
\num{6.04}; \num{1.99} vs.\ \num{2.12} with classifier-free guidance, Appendix~\ref{app:arch}; see Appendix~\ref{app:sig} for the size of this margin), every structure-agnostic objective beats the undistilled \emph{No-Distillation}
(\num{9.52}), and each reaches a given gFID in fewer training steps (Figure~\ref{fig:convergence}).
\item \emph{No grid}: can we drop correspondence entirely? Yes. In the 1D-sequence latent, the pooled objectives still beat \emph{No-Distillation}
(\emph{Pool-Align} \num{15.62} vs.\ \num{26.49}) in the same order.
\item \emph{Cross-modal}: can structure-agnostic distillation cross modalities? Yes, for two of the three objectives: with a
text teacher, \emph{Pool-Align} (\num{6.97}) and \emph{Soft-KL} (\num{8.51}) improve over
\emph{No-Distillation} (\num{9.52}), while \emph{CKA} (\num{10.41}) lands above it.
\end{enumerate}

\section{Discussion}
\label{sec:discussion}
\paragraph{Why pooling can match position-wise alignment.} We hypothesise that pooling helps
in-modality by averaging out distractors carried in the teacher's per-token features. Positional
information is one such distractor: self-supervised ViTs encode substantial positional signal in
their tokens \citep{amir2021dinovit}, and position-wise alignment copies it into the latent,
where it competes with the content the decoder must reconstruct; pooling averages it out. High-norm
artifact tokens \citep{darcet2024registers} are another such distractor. A latent PCA (Appendix~\ref{app:pca}) is consistent with this picture: the
position-wise VA-VAE latent inherits a smooth spatial gradient artefact that the pooled latents lack.

\paragraph{What transfers across modalities.} With a text teacher, \emph{Pool-Align} brings gFID
from \num{9.52} to \num{6.97}, close to its \num{5.77} with DINOv2, and \emph{Soft-KL} also
improves it, so both help in every setting. Only \emph{CKA} fails across modalities; we hypothesise
that its constraint, the full centered between-image geometry, is the strictest of the three. Our
captions largely restate each image's ImageNet class, so part of the gain may be class-level semantics. We test one text encoder and caption source, and not text-to-image generation, where a
conditioning-aligned teacher could help more.

\section{Conclusion}
\label{sec:conclusion}
We show that latent distillation can improve diffusability without requiring spatial
correspondence. Pooling representations performs as well as or slightly better than position-wise alignment for image teachers, and extends distillation to 1D latents and to a
text teacher, where \emph{Pool-Align} and \emph{Soft-KL} improve generation quality. Structure-agnostic distillation
is thus a simple, flexible alternative to position-wise alignment that leaves the latent's layout free.

\bibliographystyle{plainnat}
\bibliography{references}

\appendix
\section{Architecture and training details}\label{app:arch}

\paragraph{VA-VAE and LightningDiT.} \citet{yao2025vavae} observe that high-dimensional latents
reconstruct well but diffuse poorly, and fix this by aligning the latent to a vision foundation
model. VA-VAE is a KL-regularized convolutional VAE trained with their ``VF loss'': with
$\tilde z_j = W z_j$ the projected latent at position $j$, $g_j$ the co-located DINOv2 feature, $N$
positions per image and $[x]_+=\max(x,0)$,
\[
\mathcal{L}_{\text{VF}}
= \frac{1}{N}\sum_{j}\bigl[\,1 - m_c - \cos(\tilde z_j, g_j)\bigr]_+
+ \frac{1}{N^2}\sum_{j,k}\bigl[\,\lvert\cos(\tilde z_j,\tilde z_k)-\cos(g_j,g_k)\rvert - m_d\bigr]_+ ,
\]
with $m_c=0.5$, $m_d=0.25$, averaged over the batch: a position-wise cosine match plus a
within-image analogue of our relational objectives. \emph{Pool-Align} is the first term on pooled
descriptors without margin; our \emph{VF} baseline is this model retrained from the released code.
LightningDiT is their rectified-flow transformer prior \citep{liu2023rectified,
lipman2023flowmatching} (linear path, velocity prediction, lognormal timestep sampling, cosine
velocity loss; AdamW with $\beta_2=0.95$, clipping at $1$, batch $1024$); trained for $64$ epochs
on a $50$-epoch VA-VAE, our exact budget, it reaches gFID $2.11$, which \emph{VF} reproduces. All
arms share this pipeline; only the distillation term changes.

\paragraph{Autoencoder.} We build on VA-VAE \citep{yao2025vavae}. The 2D-grid tokenizer
is its convolutional VAE: input $256\times256$, base width $128$, channel multipliers
$(1,1,2,2,4)$ (four $2\times$ downsamples, an $f16$ reduction), two residual blocks per
stage, self-attention at $16\times16$, and a $32$-channel latent, giving a
$32\times16\times16$ posterior. The 1D-sequence tokenizer keeps this convolutional
encoder and decoder unchanged and replaces only the bottleneck by a Perceiver-style
resampler \citep{jaegle2021perceiver}: $K=32$ learned latents of dimension $128$
cross-attend to the $16\times16$ conv tokens ($2$ attention blocks, $8$ heads) to form
the token sequence, and $256$ learned position queries cross-attend back for decoding.

\paragraph{Diffusion prior.} We adopt LightningDiT-XL \citep{yao2025vavae}: $28$ blocks,
hidden size $1152$, $16$ heads, patch size $1$, with RMSNorm \citep{zhang2019rmsnorm},
SwiGLU \citep{shazeer2020glu} and RoPE \citep{su2021roformer}. For the 1D setting the $32$ latent
tokens are embedded per token with a learned $1$D positional embedding (RoPE disabled). The main
comparison (Table~\ref{tab:gfid}) uses no classifier-free guidance. For the 2D-grid setting we
additionally check guidance under VA-VAE's recipe (interval guidance
\citep{kynkaanniemi2024interval}, settings in Table~\ref{tab:hparams}, each arm at its optimum
scale): \emph{VF} reproduces VA-VAE's reported FID (\num{2.12} vs.\ \num{2.11})
and \emph{Pool-Align} stays strongest (\num{1.99}), so the pipeline is faithful and the ranking
survives guidance. This recipe amplifies the leading latent channels and does not transfer to the
1D or cross-modal latents, so we report guided FID only where it is meaningful.

\paragraph{Teachers.} The vision teacher is a frozen DINOv2 ViT-L/14
\citep{oquab2023dinov2}; images are resized to $224\times224$, giving a $16\times16$ grid
of $1024$-dimensional patch tokens (the class token is dropped). For the cross-modal teacher
we take the per-image captions from the captioned ImageNet of \citet{degeorge2025imagenett2i}
and encode each with BGE-large \citep{xiao2023bge}, precomputing one text embedding per
training image and matching against it, so this teacher carries no spatial grid. \emph{Pool-Align}'s projector is a
single learned bias-free linear map from the latent into the teacher's dimension.

\subsection{Training and evaluation hyperparameters}\label{app:hparams}
Table~\ref{tab:hparams} gives the full optimization, training and sampling configuration, shared
by every arm; only the per-method distillation settings (last block) and the latent shape differ.

\begin{table}[h]
\centering
\small
\begin{tabular}{@{}l p{0.62\linewidth}@{}}
\toprule
\multicolumn{2}{@{}l}{\textbf{Autoencoder (tokenizer), all arms}}\\
\midrule
Optimizer & Adam, $\beta=(0.5,\,0.9)$, weight decay $0$\\
Learning rate & $10^{-4}$, constant (no warmup or schedule)\\
Global batch size & $256$ ($8$ per GPU $\times$ $32$ GPUs)\\
Epochs & $50$\\
Precision & fp32\\
Gradient clipping & global norm $1.0$\\
Weight averaging (EMA) & none\\
Pixel reconstruction (L1) weight & $1.0$\\
LPIPS (VGG) perceptual weight & $1.0$\\
Reconstruction NLL & learned scalar log-variance (init $0$)\\
KL weight & $10^{-6}$\\
Discriminator & PatchGAN, $3$ layers, $64$ base channels, hinge loss\\
Discriminator start / weight & step $5001$ / $0.5\times$ adaptive\\
Distillation weight $\lambda$ & gradient-norm balanced (Appendix~\ref{app:lambda})\\
Distillation target & encoder posterior mean\\
\midrule
\multicolumn{2}{l}{\textbf{Diffusion prior (LightningDiT-XL/1)}}\\
\midrule
Optimizer & AdamW, $\beta=(0.9,\,0.95)$, weight decay $0$\\
Learning rate & $2\times10^{-4}$, constant\\
Global batch size / steps & $1024$ / $80\,000$ ($\approx 64$ epochs)\\
Precision & bf16 mixed\\
Gradient clipping & global norm $1.0$\\
Weight averaging (EMA) & $0.9999$\\
Objective & linear flow matching (velocity) $+$ cosine loss\\
Timestep sampling & logit-normal\\
Latent preprocessing & per-channel standardization\\
Class-label dropout (for CFG) & $0.1$\\
\midrule
\multicolumn{2}{l}{\textbf{Sampling and FID}}\\
\midrule
Sampler & Euler ODE, $250$ steps\\
Samples for FID & $50\,000$\\
Guidance (unguided / guided) & cfg $1$ / interval from $0.11$, timestep shift $0.3$, optimum cfg per arm\\
\midrule
\multicolumn{2}{l}{\textbf{Per-method distillation settings}}\\
\midrule
\emph{VF} (position-wise) & DINOv2 teacher, weight target $r=0.1$, distance-matrix margin $0.25$, cosine margin $0.5$\\
\emph{Pool-Align} & DINOv2 teacher, pooled cosine, weight target $r=10$, cosine margin $0$\\
\emph{CKA} & pooled unbiased linear-CKA kernel, weight target $r=10$\\
\emph{Soft-KL} & pooled kernel, softmax temperature $\tau=0.1$, weight target $r=10$\\
\bottomrule
\end{tabular}
\caption{Full training and evaluation configuration. The autoencoder trains first, then the
diffusion prior on its frozen latents. Data is ImageNet-1k at $256\times256$, normalized to
$[-1,1]$, with random-crop augmentation (no horizontal flip). The weight target $r$ is the
gradient-norm-balancing factor of Appendix~\ref{app:lambda}.}
\label{tab:hparams}
\end{table}

\subsection{VAE training loss}\label{app:vaeloss}
$\mathcal{L}_{\text{vae}}$ is the standard VA-VAE tokenizer objective, unchanged by our
distillation term. It combines an $L_1$ pixel reconstruction and a VGG-LPIPS perceptual
loss \citep{zhang2018lpips}, wrapped in a learned-variance negative log-likelihood; a KL
term pulling the posterior toward a standard normal, weighted by $10^{-6}$: at this weight it is
not a generative prior but a light regularizer that keeps the latent's scale and posterior variance
bounded, so the latent handed to the diffusion prior stays smooth and well-conditioned; and a hinge
adversarial loss from a PatchGAN discriminator \citep{isola2017pix2pix}, enabled after $5000$ steps and weighted
by $0.5$ times an adaptive factor. The adaptive factor is the ratio of the reconstruction
and adversarial gradient norms at the decoder's last layer, the standard VA-VAE balancing.

\paragraph{Distillation weight.}\label{app:lambda}
As in VA-VAE, $\lambda$ is set by gradient-norm balancing at the encoder's last layer ($r=10$ for
our objectives; VA-VAE's own $r=0.1$ for \emph{VF}, whose configuration we leave untouched and which
reproduces its reported gFID, \num{2.12} vs.\ \num{2.11}). The distillation term acts on the
posterior mean rather than the reparameterized sample, which would let the encoder inflate the
posterior variance.

\subsection{Unbiased CKA}\label{app:cka}
The \emph{CKA} objective uses the minibatch unbiased linear-CKA estimator \citep{kornblith2019cka},
$\mathrm{CKA}(S_s,S_t)=\mathrm{HSIC}(S_s,S_t)/\sqrt{\mathrm{HSIC}(S_s,S_s)\,\mathrm{HSIC}(S_t,S_t)}$,
where, with $\tilde S = S - \mathrm{diag}(S)$ the $B\times B$ kernels with zeroed diagonals (so only
between-image structure enters),
\begin{equation}
\mathrm{HSIC}(K,L) = \frac{1}{B(B-3)}\left[ \mathrm{tr}(\tilde K\tilde L)
+ \frac{(\mathbf{1}^\top\tilde K\mathbf{1})(\mathbf{1}^\top\tilde L\mathbf{1})}{(B-1)(B-2)}
- \frac{2}{B-2}\,\mathbf{1}^\top\tilde K\tilde L\mathbf{1} \right].
\end{equation}

\section{Additional results and analyses}\label{app:results}
\subsection{Significance of gFID gaps}\label{app:sig}
gFID varies with the training and sampling seed at a coefficient of variation near $1.3\%$
\citep{dufour2026fidlottery}, i.e.\ $\sim\!0.1$--$0.2$ FID at our operating points, the scale
\citet{dufour2026fidlottery} recommend as the smallest difference worth reading. Our improvements
over \emph{No-Distillation} exceed it, the smallest (cross-modal \emph{Soft-KL}, $\sim\!1$ FID) several
times over. The \emph{Pool-Align} margin over \emph{VF} (\num{0.27}, \num{0.13} with guidance) sits
just above it and comes from one run per objective, so we read \emph{Pool-Align} as on par with or slightly better than \emph{VF}.

\subsection{Latent PCA}\label{app:pca}
\begin{figure}[h]
\centering
\includegraphics[width=\linewidth]{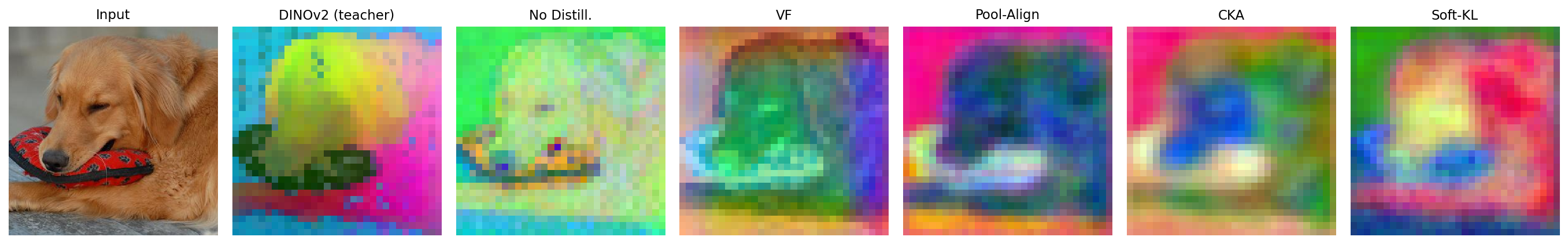}
\caption{\textbf{Position-wise alignment carries a spatial gradient; pooled latents track content.} Top-3 PCA (as
RGB) of one image's representation for the DINOv2 teacher, the undistilled latent, and each
distilled latent. The
undistilled latent is nearly featureless; the position-wise VA-VAE (\emph{VF}) latent carries a
smooth horizontal spatial gradient, whereas the pooled latents show no such gradient and instead pick out the
foreground object from the background, tracking content rather than residual information from positional embeddings in the teacher's representations.}
\label{fig:pca}
\end{figure}

\subsection{Samples}\label{app:samples}
\begin{figure}[!h]
\centering
\includegraphics[width=0.8\linewidth]{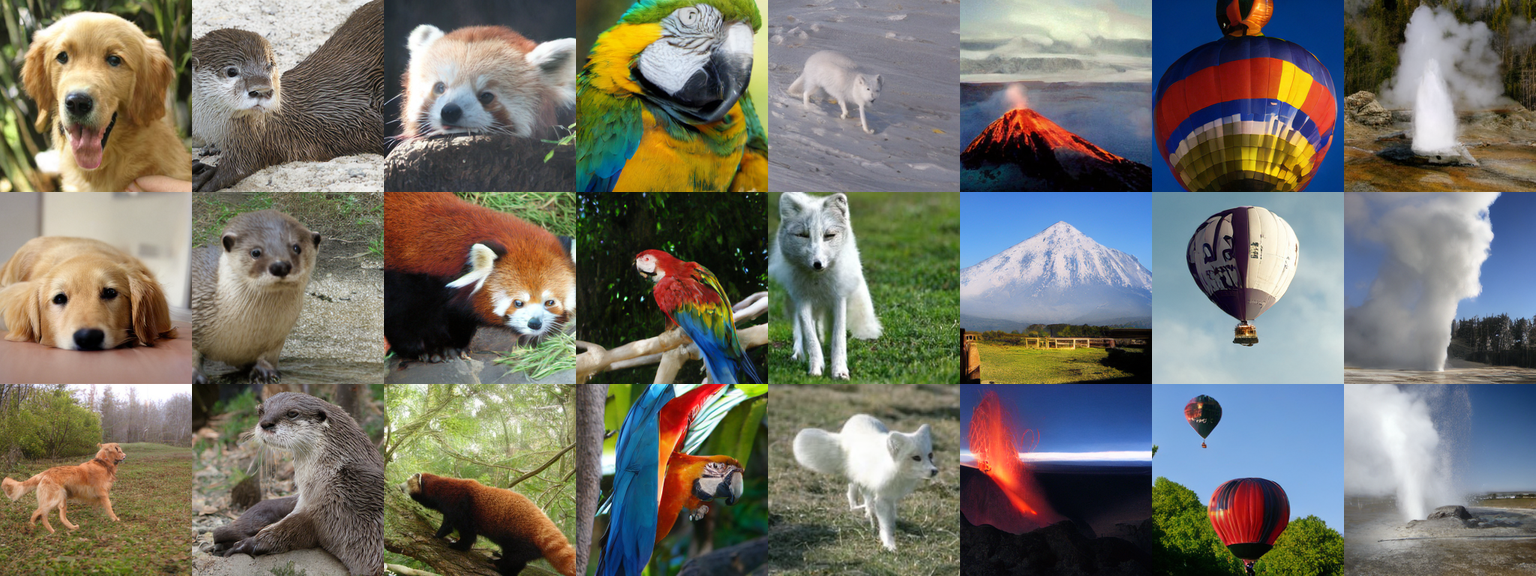}
\caption{\textbf{Samples from \emph{Pool-Align} on the 2D grid} (DINOv2 teacher,
setting~A): autoencoder trained for $50$ epochs, diffusion prior for $64$ epochs ($80$k steps);
sampled in $250$ Euler steps with classifier-free guidance (recipe in Appendix~\ref{app:arch}),
which corresponds to gFID $1.99$.}
\label{fig:samples}
\end{figure}

\end{document}